\documentclass[11pt]{article}

\usepackage[final]{acl}
\usepackage{amssymb}
\usepackage{algorithm}
\usepackage{algorithmic}
\usepackage{multirow} 
\usepackage{tabularx}
\usepackage{amsmath}
\usepackage{acronym}
\usepackage{pifont}
\usepackage[dvipsnames]{xcolor}
\usepackage{times}
\usepackage{latexsym}
\usepackage{booktabs}   %
\usepackage{pgfplots}   %
\usepackage{caption}    %
\pgfplotsset{compat=1.18}
\usepackage[T1]{fontenc}
\usepackage{listings}
\usepackage{xcolor}

\usepackage[utf8]{inputenc}

\usepackage{microtype}

\usepackage{inconsolata}

\usepackage{graphicx}

\newcolumntype{R}{>{\raggedleft\arraybackslash}X}

\title{UCVE: Graph Structured Retrieval for Compliance Validation}

\author{Emily Chen  \quad  Michelle Menachery \quad  Rahul Nenavath  \quad  Himanshu Kumar \\  \textbf{Amirfarrokh Iranitalab} \\
    Capital One, AI Foundations \\
    \texttt{\{emily.chen2, michelle.menachery, rahul.nenavath, himanshu.kumar2,} \\ \texttt{amirfarrokh.iranitalab\}}\texttt{@capitalone.com}
}

\newacro{UCVE}{Universal Content Validation Engine}
\newacro{CRG}{Contextual Requirement Graph}
\newacro{LLMs}{Large Language Models}
\newacro{LLM}{Large Language Model}
\newacro{GenAI}{Generative AI}
\newacro{RL}{Reinforcement Learning}
\newacro{GSAEPS}{Graph-Structure-Aware Entry Point Selection}
\newacro{BFS}{Breadth First Search}
\newacro{PE}{path efficiency}
\newacro{PPR}{Personalized Page Rank}

\begin{document}
\maketitle
\begin{abstract}
Structured compliance validation requires determining whether every obligation in a predefined requirement space is satisfied, violated, or unaddressed by a given document. This can be described as a coverage problem that existing retrieval and classification systems, by treating requirements as independent, are not designed to solve. We introduce \textbf{\ac{UCVE}}, a domain-agnostic framework that represents compliance requirement spaces as a \ac{CRG} and performs retrieval via spectral reranking over its reinforcement learning-weighted Laplacian. Edges between requirement nodes are learned via reinforcement learning with a traversal reward grounded in violation detection, reshaping the graph's community structure such that low-pass spectral filtering distinguishes relevant from irrelevant obligation clusters. Evaluating across three compliance domains, \ac{UCVE} averages 4.3\% matched-$K$ recall lift over flat cosine retrieval and consistently outperforms strong retrieval and graph-based baselines.

\end{abstract}
\section{Introduction}
Legal compliance review requires verifying that a document satisfies, or avoids breaching, a structured set of legal obligations. Privacy policies must disclose the data practices mandated by regimes such as the GDPR and CCPA, consumer contracts must avoid terms that are potentially unfair under the EU Unfair Contract Terms Directive (93/13/EEC) \cite{lippi2019claudette}, and Environmental, Social, and Governance (ESG) reports must address disclosure criteria set by frameworks such as the EU Corporate Sustainability Reporting Directive. As regulatory volume outpaces manual review by lawyers, compliance officers, and consumer protection authorities, automated systems that can handle the nuance of legal language are increasingly necessary \cite{chalkidis-etal-2022-lexglue}. We study this problem through the \textit{Universal Content Validation Engine} (UCVE), a framework that utilizes a Contextual Requirement Graph (CRG) for structured retrieval over a predefined space of legal obligations. The costs of error in this setting are asymmetric: an overlooked unfair term or missing mandatory disclosure can expose an organization to regulatory sanctions, litigation, or reputational harm, whereas a false alarm costs only additional human review. High recall over the obligation space is therefore the central requirement for practical legal validation systems.

Existing legal NLP approaches to compliance typically treat obligations as independent units during inference. Classification models predict clause labels in isolation, and standard retrieval methods extract evidence for individual requirements without propagating information across related obligations. Legal obligations, however, are rarely independent. Statutes and contract taxonomies are hierarchically organized, provisions cross reference one another, and the same obligation is often satisfied through varied drafting and paraphrase. Local retrieval fails to capture these semantic and hierarchical interdependencies \cite{edge2024graphrag}. For example, a data retention clause in a privacy policy frequently bears on storage limitation and third party sharing obligations, and a jurisdiction clause in terms of service commonly co-occurs with a choice of law clause. Because traditional strategies do not systematically exploit these dependencies, they often miss violations that lack direct lexical overlap with the statutory or regulatory requirement text. Furthermore, many systems are tightly coupled to a single jurisdiction or annotation schema, making them difficult to adapt as laws are amended or new regulatory regimes emerge without exhaustive reannotation \cite{zhong-etal-2020-nlp}.

To address these limitations, \ac{UCVE} models legal validation as inference over a CRG where nodes represent obligations and edges encode the semantic and hierarchical relationships between them. Following initial retrieval, the system propagates relevance signals across the graph using spectral reranking to smooth similarity scores across community structures. Edge weights are optimized via a reinforcement learning objective that rewards the recovery of violations during traversal, allowing the system to identify obligations supported by indirect or distributed evidence across clauses. Each node carries a natural language validation question grounded in its requirement text, so every verdict is traceable to a specific obligation, supporting the auditability that legal practitioners require. Because the graph is built from explicit requirement definitions rather than task specific outputs, the architecture is portable across heterogeneous legal compliance domains.

\begin{figure*}[h]
    \centering
    \includegraphics[scale=0.65]{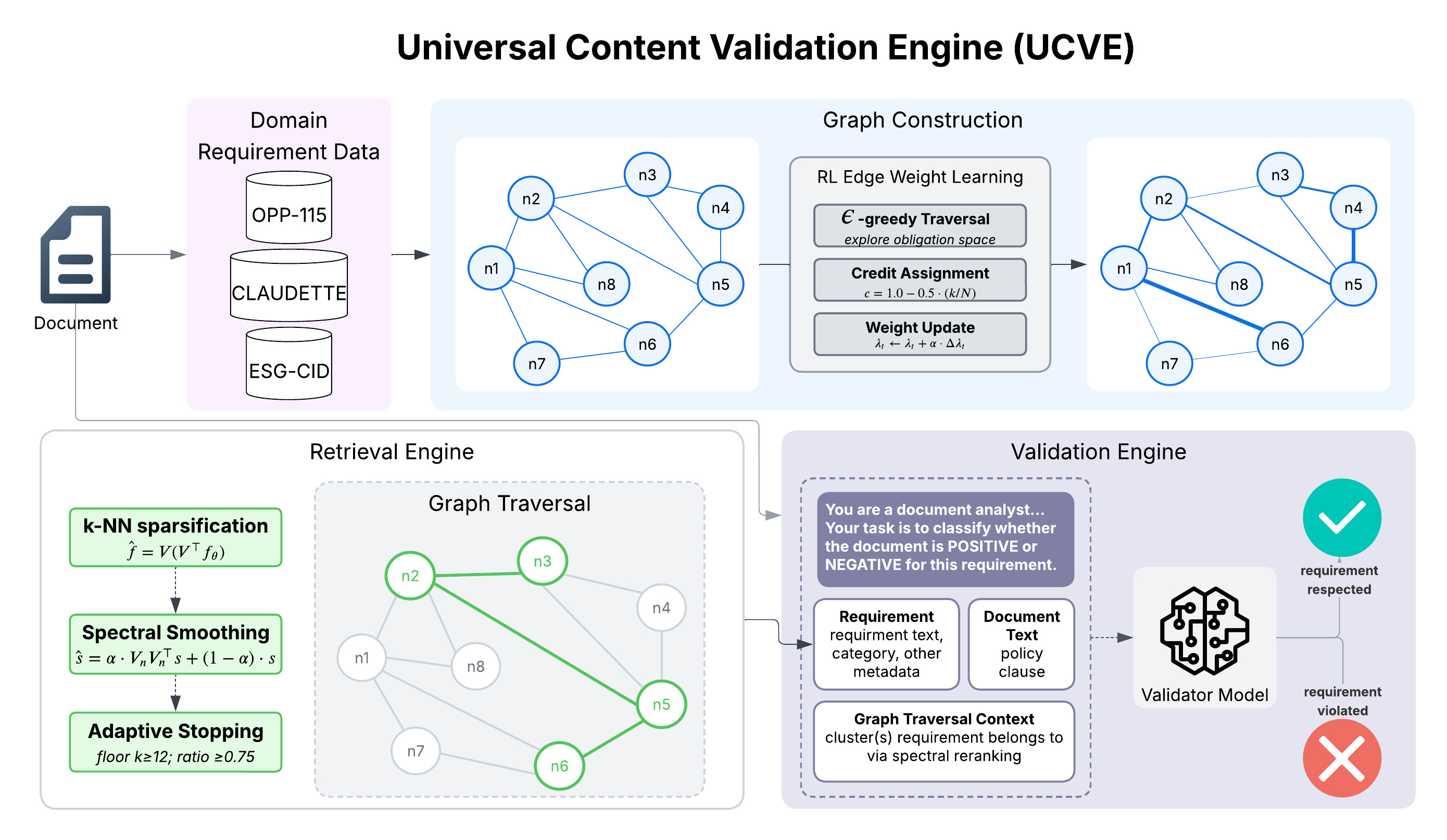}
    \caption{Universal content validation engine architecture.}
    \label{fig:ucve_architecture}
\end{figure*}

We evaluate the framework on three legal compliance benchmarks covering privacy policies \cite{wilson2016opp115}, consumer contract fairness \cite{lippi2019claudette}, and sustainability disclosures. Focusing on document level recall, the approach consistently outperforms strong retrieval and graph based baselines, particularly for obligations whose evidence spans multiple clauses. Our contributions are threefold: we formulate legal
compliance validation as structured inference over an interconnected obligation graph under asymmetric legal risk; we introduce the CRG, with RL-learned edge weights combined with spectral reranking to propagate relevance through graph structure; and we show the architecture generalizes across structurally distinct legal domains with only lightweight per-domain adaptation.

\section{Related Work}

\subsection{Legal NLP and Compliance Checking}
Early compliance automation checked extracted obligations against target documents \cite{breaux2006towards, cejas2023nlp} with subsequent systems introducing explicit graph representations for verification \cite{tauqeer2022automated}. Recent graph-based compliance systems differ in their graph provenance. PrivComp-KG \cite{garza2024privcomp} pairs LLM retrieval with a regulatory knowledge graph for privacy-policy compliance verification, whereas GraphCompliance \cite{chung2025graphcompliance} aligns a regulatory Policy Graph with an input-specific Context Graph built per document. Similarly, ComplianceNLP \cite{guo2026compliancenlp} grounds obligation extraction and policy-gap analysis in a reusablable regulatory knowledge graph. SAT-Graph RAG \cite{demartim2025satgraph} retrieves deterministically over an LRMoo-inspired temporal legal graph. Obligation Logic Graphs \cite{servantez2023computable} model contractual obligation dependencies for computable contract execution. UCVE instead treats the obligation space itself as a document-independent requirement graph, learning inter-requirement edge importance from validation outcomes.

ContractNLI \cite{koreeda-manning-2021-contractnli-dataset} checks a fixed hypothesis set checked against each document, the closest structural analogue to our task, but classifies each hypothesis independently. \cite{cejas2023nlp} and \cite{abualhaija2022automated} share this closed-set framing over curated GDPR requirements, but model no dependencies across the requirement space.

Adjacent Legal NLP benchmarks target clause identification (CUAD \cite{hendrycks2021cuad}), statutory and case retrieval (BSARD \cite{louis-spanakis-2022-statutory}; COLIEE \cite{rabelo2022overview}), general legal reasoning (LegalBench \cite{guha2023legalbench}), excerpt-level regulatory judgments (AIReg-Bench \cite{marino2025airegbench}), and violation identification in unstructured text (LegalLens \cite{bernsohn2024legallens}), query-to-corpus or per excerpt tasks distinct from UCVE's closed-set, per-obligation coverage.

\subsection{Graph-based RAG and Knowledge Graph Approaches}

Recent Graph-RAG advancements couple LLMs with Knowledge Graphs (KGs) for complex reasoning via agentic beam search \cite{sun2024think} or explicit relation planning \cite{luo2024rog}.  Surveyed Graph-RAG methods \cite{peng2025graph} overcome isolated semantic retrieval through hierarchical
communities \cite{edge2024graphrag}, Maximum Spanning Trees \cite{zhu-etal-2025-knowledge}, textual subgraph retrieval \cite{hu2025grag}, and lexical proposition--entity graphs \cite{ghassel2025hierarchical}. To operationalize traversal, approaches use Personalized PageRank \cite{2024hipporag}, document-graph reranking \cite{dong2024don}, graph neural message passing over KG entities \cite{mavromatis2024gnn}, or map unstructured text to deterministic policy graphs \cite{chung2025graphcompliance}. However, these rely heavily on static semantic connectivity, pre-existing schemas, or heuristic traversals. UCVE diverges by optimizing structural metadata edges via RL, enabling the policy to explicitly learn domain-specific co-relevance patterns for high-recall obligation recovery.

\subsection{Reinforcement Learned Edges}
RL is widely used in multi-hop KG reasoning to learn traversal policies, often framed as sequential decision-making \cite{xiong2017deeppath} \cite{das2018go}. To address incomplete KGs and spurious trajectories, subsequent walk-based models employ knowledge-based reward shaping and action dropout \cite{lin2018multi}. At the structural level, methods like Iterative Deep Graph Learning continuously optimize both the adjacency matrix and GNN parameters \cite{chen2020iterative}. Closer still, several systems use RL to shape graph structure directly: Goal-Directed Graph Construction learns
  graph-construction and improvement policies with RL and GNNs \cite{darvariu2021goal}, AutoGraph-R1 trains an LLM graph constructor from task-derived rewards \cite{tsang2026autograph}, and GraphRAG-R1 applies process-constrained RL to refine multi-hop graph retrieval \cite{yu2026graphrag}. Rather than learning navigational pathways or regenerating entire graph topologies from continuous embeddings, UCVE uses RL strictly to weight discrete, domain-specific structural metadata (e.g., hierarchical links). By rewarding the recovery of missed obligations during $\epsilon$-greedy traversal, UCVE dynamically discovers the structural relationships that genuinely encode co-relevance.

\subsection{Spectral Graph Theory}

Spectral graph theory formally propagates signals across networks, capturing global node importance via the dominant eigenvector of transition matrices \cite{page1999pagerank}. Graph Signal Processing extends this by framing retrieval as linear filtering over a spectral domain; projecting queries onto a low-rank Laplacian eigenbasis allows low-pass filters to smooth relevance signals and accelerate manifold traversal \cite{ortega2018graph}. Within spectral ranking, Iterated Graph Laplacians cast manifold ranking as Laplacian regularization \cite{zhou2011iterated}, Fast Spectral Ranking recasts manifold retrieval as frequency-domain linear filtering \cite{iscen2018fast}, and Query-Aware Flow Diffusion adapts propagation via query-aligned edge flows \cite{zhou2026query}. Rather than flat similarities or sequential random walks, UCVE projects traversal scores onto the bottom
eigenvectors of an RL-weighted, sparsified graph Laplacian. This low-pass filter smooths signals across learned requirement communities and suppresses high-frequency noise, lifting lexically dissimilar but structurally relevant obligations to maximize recall.

\section{Methods}
\ac{UCVE} is a two-stage framework for requirement validation over arbitrary document collections with a given domain. The high level pipeline architecture is shown in Figure \ref{fig:ucve_architecture}.

\subsection{Context Requirement Graph}
We introduce a \ac{CRG}, a weighted, undirected graph $G=(V, E, w)$ over a domain's requirement space. The \ac{CRG} is created separately for each domain, as these graphs are anticipated to be used independently. The graph is created independent of any target document and encodes domain specific metadata explicitly in the graph structure. Each node $v \in V$ represents a single requirement drawn from a domain specific requirement set. Each edge $e \in E$ encodes a relationship between two requirements (semantic similarity of requirement descriptions or structural metadata) with a composite weight $w(e)$ aggregating across multiple typed edge contributions. The graph is constructed once per dataset, independent of any target documents. 

\subsubsection{Node Definition}
Each node $v \in V$ stores a unique identifier, domain indicator, requirement category label and a natural language \texttt{requirement} text. To ensure that validation decisions are interpretable and auditable, each node also contains a natural-language validation question derived from its stored requirement and description. Nodes also include positive and negative class definitions that are instantiated programmatically: a positive label indicates that the document text contains language exhibiting, satisfying, or relating to the requirement, while a negative label indicates the absence of such language. This framing converts the per-node classification task into a grounded binary judgment that requires no dataset-specific prompt engineering, where the same template applies uniformly across all domains and requirement types.

\subsubsection{Embedding}
Node embeddings contribute to edge construction and query-node similarity scoring during traversal. We evaluated node requirement texts under two encoders, \texttt{all-mpnet-base-v2} \cite{reimers-2019-sentence-bert}, a widely used domain-general sentence encoder, and \texttt{bge-large-en-v1.5} \cite{xiao2024cpack}, a more recent encoder from a different training lineage. We deliberately prefer a domain-general encoder rather than a domain adapted model to ensure that the \ac{CRG} has the ability to capture relationships that stem from requirement semantics and graph structure rather than a pattern specific to the wording of a given domain. We adapt \texttt{all-mpnet-base-v2} for the main results, as it achieves higher or comparable recall across all three domains in our evaluation (Appendix \ref{app:encoder-sensitivity}). 

\subsubsection{Semantic Edge Construction}
The graph is initialized by connecting nodes whose requirement texts are semantically similar, establishing a backbone that reflects the latent structure of the requirement space. For every pair of nodes ($v_i , v_j$), we compute cosine similarity over their L2-normalized embeddings. 

\begin{equation}
sim(v_i , v_j) = e_i \cdot e_j
\end{equation}

where $e_i$ is the unit-length embedding of node $v_i$. An undirected edge is created if $sim(v_i , v_j) \geq \tau$, with threshold $\tau=0.5$, and edge weight equal to cosine similarity value. This threshold admits semantically related requirements while excluding illegitimate connections between unrelated obligation categories.   

\subsubsection{Structural Edge Enrichment} \label{sec:edges}
Semantic similarity alone cannot encode relational structure that domain experts encode in the datasets themselves such as hierarchical requirement structures. We augment the graph with six additional edge types derived from dataset ingrained metadata, producing a multi-relational graph with seven edge types in total.

Each edge holds an array of typed values, where each value is a tuple \texttt{(type, raw weight, learned weight)}. The \texttt{raw weight} encodes the intrinsic strength of the relationship: for a cosine similarity typed value, this would be a value $\in [0.5, 1.0]$. For the rest of the metadata types, it is a binary indicator (1.0 if both nodes share that property). The \texttt{learned weight} is a scalar multiplier that get computed per type through \ac{RL} (Section \ref{sec:rl-edges}), initialized to 1.0 for cosine similarity and 0.0 for all metadata types. The composite edge weight aggregated across all active types:

\begin{equation}
w(e)= \sum_{t\in type(e)} raw\_wght_t \times learned\_wght_t 
\end{equation}

Metadata edges can only increase the total weight, encoding the intuition that nodes related on multiple dimensions are more likely to co-occur as relevant requirements for the same document. The structural edge types instantiated for a given domain are determined entirely by the metadata fields present in that domain's dataset, requirement no changes to the framework's core architecture.

\subsection{RL Edge Weight Learning} \label{sec:rl-edges}

The structural edge types introduced in Section~\ref{sec:edges}
augment the graph with domain-specific relational signals whose relative importance varies across domains. Rather than hand-tuning edge type weights per domain, we learn a weight vector $\lambda = [\lambda_1, \ldots, \lambda_6]$ over the six learnable metadata edge types via reinforcement learning, with the cosine similarity weight fixed at 1.0 throughout. The objective is to maximize retrieval recall over the training split — the fraction of ground truth requirement nodes recovered within a fixed traversal budget.

\paragraph{Traversal and credit assignment.} Training uses an $\epsilon$-greedy breadth-first traversal over the domain graph. For each training document, traversal begins at the node with highest query similarity and proceeds for $B{=}15$ steps, scoring unvisited neighbors by a weighted combination of query similarity and current edge weight (Equation~\ref{eq:traversal-score}). At each step, credit is assigned to the traversed edge based on the destination node's ground-truth status: edges leading to a true violation receive credit that decays with step index $k$ (rewarding earlier discovery over later discovery), edges leading to a node within three hops of a violation receive a small positive credit, and all other edges receive a small negative credit. This credit is distributed across a traversed edge's active metadata types in proportion to their relative contribution to that edge's composite weight, and each type's weight is updated via gradient ascent, clamped to $[0.0, 3.0]$ to prevent runaway growth while preserving the constraint that metadata edges only ever increase relative importance over the cosine-similarity backbone. Training runs for $T{=}30$ epochs with learning rate $\alpha{=}0.05$ and $\epsilon$ decaying from $0.15$ to $0.02$. Full credit assignment and update equations are given in Appendix~\ref{app:rl-details} and Algorithm~\ref{alg:rl}.

\paragraph{$\epsilon$-greedy BFS.} UCVE uses RL only to weight a small, fixed set of typed metadata edges rather than learn a navigational policy. This is a
substantially simpler credit-assignment problem than multi-hop knowledge graph reasoning. $\epsilon$-greedy suits this setting: obligation graphs are small and bounded (24-82 nodes), making policy-gradient machinery unnecessary. Breadth-first expansion naturally promotes the coverage our recall-centric objective rewards \ref{sec:metrics} and $\epsilon$-greedy expansion guarantees credit reaches every edge type, which purely greedy traversal would not. Policy-gradient traversal becomes the natural choice only at substantially larger obligation spaces where exhaustive BFS is intractable (Limitations).

\subsection{Requirement Retrieval Engine}
At inference time, the \ac{CRG} is applied to a given document through a three-step procedure. First, entry point selection identifies the most structurally promising nodes from which to initialize traversal. Next, traversal explores the graph from the entry points to retrieve a set of requirement nodes. Lastly, spectral re-ranking refines the candidate set using the community structure encoded in the \ac{RL}-weighted graph. Together, these steps produce a ranked set of requirement nodes representing the obligations most relevant to the document.

\subsubsection{Traversal}

Given a query document $q$, \ac{UCVE} scores all same-domain
requirement nodes in the \ac{CRG} using spectral reranking over the
\ac{RL}-weighted graph. Rather than sequentially traversing the graph
from entry points, spectral reranking operates globally: it smooths
cosine similarity scores across the graph's community structure, lifting nodes whose wording diverges from the query but whose spectral neighborhood is relevant, and suppressing nodes with high surface similarity that sit in irrelevant clusters. Alternative traversal strategies including entry-point selection via \ac{GSAEPS}, adaptive \ac{BFS}, and simultaneous graph scoring are evaluated in Appendix~\ref{app:traversal}.

\subsubsection{Spectral Reranking} \label{sec:spectral-reranking}

\textbf{Sparsification.} The \ac{RL}-weighted graph is fully dense by construction: every node pair carries a non-zero composite weight. Standard graph propagation methods such as heat diffusion, personalized PageRank, embedding propagation, degenerate toward flat cosine similarity ranking, as every node is a neighbor of nearly every other, leaving no discriminative community signal. We address this through $k$-NN sparsification, retaining only the $k$ strongest edges per node. The choice of $k$ trades off two failure modes: too small a $k$ risks disconnecting the graph, while too large a $k$ approaches the dense graph and reintroduces the degeneracy sparsification is meant to avoid; we select $k$ per domain via sensitivity analysis on validation data (Section~\ref{sec:robustness}, Figure~\ref{fig:failed_constraints}).

\textbf{Spectral Decomposition.} Over the sparsified graph, we compute the unnormalized graph Laplacian $L = D - W$, where $D$ is the degree matrix and $W$ is the adjacency matrix of \ac{RL}-weighted edge weights. Eigenvectors corresponding to small eigenvalues vary slowly across the graph and capture large-scale community structure, while those corresponding to large eigenvalues capture fine-grained within-cluster variation \cite{von2007spectralclustering, chung1997spectralgraph}. We retain the bottom $n$ eigenvectors $V_n \in \mathbb{R}^{N \times n}$ as a smoothness basis for reranking, where $n$ and the sparsification parameter $k$ are selected per domain on validation data to reflect differences in the natural community structure of each requirement graph. Setting $n$ too large risks overfitting to fine-grained topology; setting it too small discards useful inter-community signal.

\textbf{Projection and Reranking.} Given the vector of cosine similarities
$s \in \mathbb{R}^N$ over all same-domain nodes, we project into the
subspace spanned by $V_n$:
\begin{equation}
    \hat{s} = \alpha \cdot V_n V_n^\top s + (1-\alpha) \cdot s,
    \quad \alpha = 0.8
\end{equation}
The projection $V_n V_n^\top s$ is a low-pass filter in the graph frequency
domain, retaining community-level similarity signal and attenuating
high-frequency within-cluster noise \cite{von2007spectralclustering}. This
has two complementary effects: (1) it lifts requirement nodes with low
cosine similarity when their spectral neighbors score highly, recovering
obligations whose wording does not closely match the query but whose
community is relevant; (2) it suppresses false positives with high cosine
similarity that sit in spectral clusters where most nodes score low. The
blend weight $\alpha{=}0.8$ prioritizes spectral community structure while
retaining direct semantic signal; setting $\alpha{=}1.0$ discards raw
similarity entirely and was found to slightly degrade performance on queries
where a single node is distinctly relevant. Nodes are reranked by $\hat{s}$
and returned up to an adaptive budget controlled by a similarity threshold
of 0.75.

\textbf{\ac{RL} Edge Weights.} The utility of spectral projection
depends on whether the sparse graph's community structure encodes
ground-truth co-relevance patterns. On a cosine similarity graph,
communities reflect embedding similarity clusters, which may not align
with how requirements co-occur in documents. On the \ac{RL}-weighted graph,
edges trained to predict co-relevance reshape the Laplacian's community
structure such that spectral clusters correspond more closely to sets of
requirements that genuinely co-occur --- the key mechanism underlying the
conditional dependence between \ac{RL} edge learning and spectral reranking
demonstrated empirically in Section~\ref{sec:results}.

\section{Results} \label{sec:results}

\begin{table*}[t]
\centering
\begin{tabular*}{\textwidth}{@{\extracolsep{\fill}} lcccccccccc @{}}
\toprule
& \multicolumn{3}{c}{OPP-115} & \multicolumn{3}{c}{CLAUDETTE} &
\multicolumn{3}{c}{ESRS} \\
\cmidrule(lr){2-4} \cmidrule(lr){5-7} \cmidrule(lr){8-10}
Method & Recall & Precision & $K$ & Recall & Precision & $K$ & Recall & Precision & $K$ \\
\midrule
BM25 & 0.649 & 0.163 & 14.0 & 0.732 & 0.104 & 16.0 & 0.4902 & 0.0640
& 16.9 \\
Flat cosine & 0.748 & 0.187 & 14.0 & 0.806 & 0.112 & 16.0 & 0.7227
& 0.0967 & 16.9 \\
Cross-encoder & 0.781 & 0.197 & 14.0 & 0.825 & 0.113 & 16.0 & 0.465
& 0.0590 & 16.9 \\
PPR & 0.583 & \textbf{0.201} & 14.0 & 0.731 & 0.106 & 16.0 & 0.5919
& 0.0825 & 16.9 \\
Leiden & 0.766 & 0.169 & 14.0 & 0.841 & \textbf{0.119} & 16.0 & 0.6745
& 0.0908 & 16.9 \\
Heat kernel & 0.747 & 0.187 & 14.0 & 0.817 & 0.114 & 16.0 & 0.7442
& 0.1000 & 16.9 \\
\midrule
CRG + Spectral & \textbf{0.792} & 0.185 & 14.2 & \textbf{0.860}
& 0.118 & 16.4 & \textbf{0.7551} & \textbf{0.1009} & 16.9 \\
\bottomrule
\end{tabular*}
\vspace{2pt}
\caption{Retrieval results at matched-$K$ (OPP-115 $K{=}14$, CLAUDETTE $K{=}16$, ESRS $K{=}17$). PPR = Personalized PageRank; Leiden = GraphRAG community detection. }
\label{tab:main-results}
\end{table*}

\subsection{Experimental Design}
\subsubsection{Datasets}
We instantiate the \ac{CRG} over three domains in the evaluation: privacy policy compliance, using OPP-115 \cite{opp115-paper} and consumer fairness detection using CLAUDETTE \cite{lippi2019claudette}, and sustainability reports using ESRS \cite{ahmed2025enhancing}. The OPP-115 dataset contains a two-tiered hierarchy of requirements with 10 category level requirements representing top-level data practice categories (i.e. First Party Collection/Use, Data Retention) and 14 sub-attribute requirements presenting more granular aspects of the top level categories (i.e. personal information type cookies is under the top level First Party Collection/Use node). The  CLAUDETTE dataset contains 53 requirements. Each CLAUDETTE node also contains dataset specific metadata: a \texttt{fairness tier} which marks if the requirement is clearly or potentially unfair and a \texttt{kb tag} field which groups semantically related requirements. Nodes are filtered to those appearing in at least 5 training documents to ensure sufficient annotation coverage. ESRS contains 82 sustainability disclosure requirement nodes drawn from the European Sustainability Reporting Standards subset of the ESRS-CID retrieval benchmark, covering disclosure requirements across environmental, social, and
governance topics. This resulting base graphs contains 577 edges connecting CLAUDETTE nodes and 148 connecting OPP-115 nodes and 595 connecting ESRS-CID (ESRS) nodes. 

All three graphs are dense (53.6\%, 99.8\% and 100\% edge density for OPP-115, CLAUDETTE and ESRS respectively).

\subsubsection{Metrics} \label{sec:metrics}
We report recall as our primary metric: the fraction of ground truth requirement nodes recovered by the system. In compliance validation, error costs are asymmetric. A requirement missed at retrieval is never passed to the validator, so a violated obligation is silently treated as addressed, a false assurance that can carry regulatory or legal consequences. An irrelevant requirement retrieved by mistake, by contrast, is cheaply rejected at the validation stage. Retrieval recall therefore sets the ceiling on end-to-end violation detection (Table~\ref{tab:validation}). We additionally report Precision@$K$, the fraction of retrieved nodes that are ground truth violations, capturing retrieval quality within the visit budget. 

The input granularity per dataset varies, reflecting the natural unit of each corpus. For OPP-115, the input is a full privacy policy document. For CLAUDETTE, the input is a single Terms of Service clause and for ESRS, the input is a single page from a sustainability report. In all cases, the system receives the input text and must recover which requirement nodes are violated. 

All graph-based methods are evaluated at their natural stopping criterion. Because methods differ in how many requirements they return per input, raw recall would conflate ranking quality with retrieval budget: a method can raise recall simply by returning more candidates. We therefore evaluate baselines at matched-$K$, where each baseline retrieves exactly $K$ nodes and $K$ equals the average number of nodes retrieved by the full \ac{UCVE} system on that dataset ($K{=}14$ for OPP-115, $K{=}16$ for CLAUDETTE, $K{=}17$ for ESRS). Recall differences thus reflect how well each method orders the requirement space rather than how many nodes it returns.

We adopt document level recall rather than clause level classification metrics reported by prior work on CLAUDETTE. We compare against six retrieval baselines at matched-$K$. BM25 \cite{robertson2009bm25} ranks requirement nodes by lexical overlap with the query document. Flat cosine ranks nodes by cosine similarity between the query and node embeddings. Cross-encoder reranks nodes using \texttt{ms-marco-MiniLM-L-6-v2} \cite{reimers-2019-sentence-bert}, a pairwise relevance model that jointly encodes the query document and candidate requirement text. \ac{PPR} runs random walks on the \ac{RL}-weighted graph seeded from the highest cosine similarity node, following HippoRAG \cite{2024hipporag}. Heat Kernel Diffusion propagates query-weighted scores over the \ac{RL}-weighted graph using the unnormalized Laplacian with diffusion time $t{=}0.5$. Community Retrieval adapts the community-detection component of GraphRAG \cite{edge2024graphrag} to the node retrieval setting via Louvain clustering, returning all nodes in the most query-relevant community.

As a robustness check on the matched-K protocol, we additionally compare each baseline under its own independently tuned adaptive stopping behavior rather than a fixed retrieval budget (Appendix~\ref{app:adaptive-stopping}). CRG + Spectral significantly outperforms flat cosine, heat-kernel diffusion, and a cross-encoder reranker on OPP-115, is statistically comparable to all three on CLAUDETTE, and outperforms flat cosine and the cross-encoder by a wide margin on ESRS, with a small and unconfirmed gap to heat-kernel diffusion. The cross-encoder's advantage narrows sharply with graph size, requiring $2.6\times$ its matched-K budget on ESRS to reach CRG + Spectral's recall. 

\subsection{Retrieval Performance}

CRG + Spectral achieves the highest recall across all three domains (Table \ref{tab:main-results}), outperforming the strongest non-spectral baseline (Leiden) by $+2.6\%$ on OPP-115 and $+1.9\%$ on CLAUDETTE, and outperforming flat cosine by $+3.2-5.4\%$ on all three domains. This demonstrates that the graph structure encodes retrieval-relevant information that embedding similarity alone cannot capture. A document derived alternative to CRG's pre-built obligation graph, evaluated in Appendix \ref{app: graph-document}, shows that document-extraction-based graph construction is unstable across domains, alternating between severe precision and recall collapse. 

ESRS, the largest and most structurally distinct domain (82 nodes, 1818 queries), demonstrates that \ac{CRG} + Spectral generalizes beyond the two primary evaluation domains even where RL edge learning does not converge (Section \ref{sec:graph-construction}). Notably, Community Retrieval (Leiden), one of the strongest non-spectral baselines on OPP-115 and CLAUDETTE, fell behind with\ac{CRG} + Spectral on ESRS (0.675 vs 0.755, Table \ref{tab:main-results}). This suggests that discrete community assignments become less reliable as the requirement space grows. Spectral reranking is robust to this degredation because it operates at node level rather than assuming nodes are fixed to clusters, smoothing score continuously across the graph Laplacian eigenbasis. 

\subsection{Graph Structure} \label{sec:graph-construction}

The recall gains over flat cosine reflect better ranking rather than visiting more nodes. Table~\ref{tab:2x2-ablation} isolates the contribution of \ac{RL} edge weight learning and spectral reranking as independent components, revealing a conditional dependency between the two.

The final learned weights (Table \ref{tab:edge_weights}) reveal a structural asymmetry across domains. OPP-115 assigns meaningful weight to hierarchical metadata,  \texttt{same\_category} (0.549) and \texttt{same\_sub\_attribute} (0.556), reflecting that sub-attribute nodes nested under the same parent category commonly co-occur as relevant in the same documents. In CLAUDETTE, metadata weights are near-zero across all types (0.037--0.041), reflecting that at high edge density the metadata structure adds minimal information beyond cosine similarity. In ESRS, category is weighted highest metadata weight (0.500) assigned as a fixed boost rather than RL-learned due to near-zero training signal, suggesting that requirements within the same ESG pillar frequently co-occur as relevant.

\begin{table}[h]              
  \centering                                                                                               
  \small                                                                                                 
  \begin{tabular}{|l|c|c|c|}    
  \hline
  \textbf{Edge Type} & \textbf{OPP-115} & \textbf{CLA} & \textbf{ESRS} \\ \hline
  \texttt{cosine\_sim}           & 1.000 & 1.000 & 1.000 \\
  \hline
  \texttt{same\_category}        & 0.549                & N/A                  & 0.500 \\
  \hline
  \texttt{same\_sub\_attribute}  & 0.556                & N/A                  & N/A                  \\
  \hline
  \texttt{same\_fairness\_tier}  & N/A                  & 0.037                & N/A                  \\
  \hline
  \texttt{same\_kb\_tag}         & N/A                  & 0.041                & N/A                  \\
  \hline
  \texttt{same\_parent\_cat}     & N/A                  & 0.040                & N/A                  \\
  \hline
  \texttt{same\_framework}       & N/A                  & N/A                  & 0.086      \\
  \hline
  \end{tabular}
  \caption{Learned edge weights per dataset. }
  \label{tab:edge_weights}
  \end{table}

Spectral reranking on the \ac{RL}-weighted graph is the only condition that improves recall on a matched-$K$ basis, achieving $+3.4\%$ lift on OPP-115, $+4.1\%$ on CLAUDETTE and $3.2\%$ on ESRS. Neither component alone is sufficient: spectral reranking on the cosine-only graph yields modest matched-$K$ lift on OPP-115 and only $+3.5\%$ on CLAUDETTE, while \ac{BFS} on the \ac{RL}-weighted graph improves over flat cosine (0.710 OPP-115, 0.806 CLAUDETTE) but underperforms spectral reranking by $-8.2\%$ and $-5.4\%$ respectively at matched-$K$.

This interaction is most evident on OPP-115, where zeroing \ac{RL} metadata weights changes 8 edges in the $k$-NN ($k{=}3$) sparsified graph --- sufficient to disrupt the spectral community structure that enables the low-pass filter to distinguish relevant from irrelevant requirement clusters. On CLAUDETTE, metadata weights are individually small but collectively meaningful: removing all metadata produces 54 topology differences and a consistent $-1.2\%$ recall loss. On ESRS, removing all metadata collectively costs $2.6\%$ recall (Table \ref{tab:edge-ablation-delta}). Together, these results confirm that \ac{RL} edge weight learning and spectral reranking are load-bearing components whose value is conditional on each other: \ac{RL} edges create the community structure that spectral projection exploits, and spectral projection is the mechanism by which that structure translates into retrieval gains. 

\subsection{Robustness} \label{sec:robustness}

CRG produces consistent gains across structurally distinct domains using hyperparameters selected per domain on validation data. OPP-115 and CLAUDETTE share a common configuration ($k{=}3$, $n{=}5$, $\alpha{=}0.8$), reflecting their smaller requirement spaces (24 and 53 nodes respectively) and well-separated community structure (Appendix \ref{app:spectral-communities}). ESRS, with 82 nodes and a denser graph, requires a wider neighborhood and richer spectral basis ($k{=}5$, $n{=}12$, $\alpha{=}0.5$) (Figure \ref{fig:failed_constraints}). The lower blend weight on ESRS reflects a greater reliance on raw cosine similarity, consistent with the observation that spectral smoothing risks over-aggregating signal in denser graphs. All three domains are well-served by the same framework with only different local hyperparameter adjustments needed. 

Sensitivity analysis over $k$-NN sparsification ($k \in [1,10]$), eigenvector count ($n \in [2,15]$), and spectral blend weight ($\alpha \in [0,1]$) confirms that the chosen configurations are stable (Figure~\ref{fig:failed_constraints}). For OPP-115 and CLAUDETTE, $k{=}3$ sparsification is optimal with sharp degradation at $k{=}1$ and gradual degradation beyond $k{=}3$. The blend weight $\alpha{=}0.8$ falls within a stable plateau spanning 0.7 to 0.9 on both datasets. For ESRS, the wider $k{=}5$ neighborhood is optimal, consistent with the larger and denser node set requiring more structural context per node.

\subsection{Full Pipeline Validation}

To evaluate the full \ac{UCVE} pipeline, we pass retrieved requirement nodes to an LLM validator that produces binary compliance verdicts. Non-retrieved nodes are treated as negative, coupling retrieval and validation quality into a single end-to-end recall metric. We evaluate \texttt{Llama-3.3-70B} \cite{Dubeyetal2024Llama3} under two validation strategies: a baseline prompt providing only the requirement text and document and a persona-calibrated prompt enriched with traversal metadata and spectral cluster membership derived from the \ac{CRG}. Results for Llama-3.1-8B are reported in Table~\ref{tab:validation-8b}.

Table~\ref{tab:validation} reports end-to-end precision, recall, and gap from retrieval ceiling. The difference between retrieval recall and recall of the full pipeline varies across domains: about 25\% on OPP-115, 4-5\% on CLAUDETTE and 23\% on ESRS. This pattern reflects a structural difference in validation difficulty across domains. CLAUDETTE's short, self-contained clauses are easily classified once retrieved because the task involves a shorter, more straightforward passage. OPP-115 and ESRS involve longer documents where evidence may be distributed throughout, making this a more difficult validation task. On ESRS, precision collapses to 0.204, attributable to class imbalance ($\sim$3\% positive rate), that causes the validator to default towards positive prediction. This aggregate figure masks substantial variation across the thirteen ESRS pillars (Table \ref{tab:esrs-pillars}). Pillars with concrete, bounded disclosure language, such as business\_conduct and pollution, achieve precision usable in practical compliance workflows, while collapse concentrates in conceptually broad
pillars whose requirements span diffuse evidence, such as affected\_communities and strategy. This indicates ESRS precision collapse is a validation difficulty correlated with requirement breadth rather than a uniform failure; pillar-specific threshold calibration or abstention is a natural mitigation we leave to future work.

\begin{table}[t]
\centering
\small
\setlength{\tabcolsep}{3pt}
\begin{tabularx}{\columnwidth}{@{} l X rrr @{}}
\toprule
\textbf{Dataset} & \textbf{Strategy} & \textbf{Prec} & \textbf{Rec} &
\textbf{$\Delta$Rec$\uparrow$\textsuperscript{\dag}} \\
\midrule
\multirow{4}{*}{OPP-115}
& Retrieval ceiling & --- & .792 & --- \\
\cmidrule(lr){2-5}
& Baseline & .871 & .535 & $-$.257 \\
& Persona & \textbf{.876} & \textbf{.543} & $-$.249 \\
\midrule
\multirow{4}{*}{CLA.}
& Retrieval ceiling & --- & .860 & --- \\
\cmidrule(lr){2-5}
& Baseline & .730 & \textbf{.817} & $-$.043 \\
& Persona & \textbf{.732} & .803 & $-$.057 \\
\midrule
\multirow{4}{*}{ESRS}
& Retrieval ceiling & --- & .755 & --- \\
\cmidrule(lr){2-5}
& Baseline & \textbf{.204} & .527 & $-$.228 \\
& Persona & .198 & \textbf{.551} & \textbf{$-$.204} \\
\bottomrule
\end{tabularx}

\vspace{4pt}
\begin{minipage}{\columnwidth}
\scriptsize
\textsuperscript{\dag}$\Delta$Rec$\uparrow$ is the gap between strategy
recall and the retrieval ceiling. Negative values indicate recall below
the ceiling; closer to zero is better. The gap reflects \emph{validation
errors} on retrieved nodes rather than \emph{retrieval failure}.
\end{minipage}

\caption{End-to-end validation performance with graph-derived context
from the \ac{CRG} using Llama-3.3-70B.}
\label{tab:validation}
\end{table}

\section{Conclusion}
Compliance validation requires retrieving relevant evidence across a structured requirement space without treating obligations as independent. We
introduced UCVE, which reframes this as structured retrieval over a Contextual Requirement Graph (CRG) whose edges encode RL-learned co-relevance among obligations. Our central finding is that RL edge weight learning and spectral reranking are conditionally dependent: RL edges reshape the graph Laplacian's community structure so that low-pass spectral filtering distinguishes relevant from irrelevant requirement clusters, and neither component alone achieves the same gains. Spectral reranking generalizes across all three domains without architectural changes, and outperforms retrieval and graph-based baselines under a document-level recall protocol that, to our knowledge, no prior compliance validation system has adopted. More broadly, this suggests spectral graph methods' retrieval utility is a function of how well edge weights encode task-relevant co-occurrence structure, not graph topology alone — a principle applicable wherever requirements, concepts, or entities exhibit learnable co-relevance that surface-level
similarity cannot capture. Extending to multi-document settings, where obligations must be satisfied across a corpus rather than a single document, is a natural next step.

\section*{Limitations}

CRG exhibits poor path efficiency on dense requirement graphs. Spectral reranking, by projecting node scores through the graph Laplacian eigenbasis, considers a large portion of the node set during scoring --- on OPP-115,  CLAUDETTE and ESRS this yields path efficiency scores of approximately 0.18, 0.12 and 0.10 respectively. However, recall\_positive remains substantially above flat cosine retrieval on these domains, and spectral reranking performance is sensitive to hyperparameter configuration ($k$-NN neighborhood size, eigenvector count, and reranking weight), with tuned configurations yielding up to 4.37\% additional lift. We interpret poor path efficiency on dense graphs as a meaningful signal about domain difficulty, and note that systematic hyperparameter search represents a direct path to further recall gains.

The three evaluation domains span 24 to 82 requirement nodes, which are relatively small requirement spaces. Our results already reveal a scalability signal: ESRS, the largest and densest graph, has a spectral signal that has a lower significance ($\lambda_2 = 0.04$) than OPP-115 ($\lambda_2 = 0.19$), and spectral reranking yields correspondingly smaller gains. This degradation has a compounding cause: larger requirement spaces require more training signal for \ac{RL} edge weight learning to converge, but larger compliance corpora tend to have sparser positive labels per node: the same class imbalance that challenges the validation stage also weakens the \ac{RL} reward signal during graph construction. On ESRS, this manifests as near-zero signal during graph construction, requiring a fixed category boost in place of data-driven edge learning. In practice, many compliance settings such as internal policy validation, model governance documentation, and privacy policy auditing operate over requirement spaces of roughly 50--150 nodes, within or near the range we evaluate; scaling beyond this range is the setting where the policy-gradient traversal discussed in Section~\ref{sec:rl-edges} would
become the more natural choice over exhaustive $\epsilon$-greedy BFS. The current CRG relies on domain experts and dataset-provided metadata schemas to define the requirement space; automating obligation graph construction directly from regulatory text, rather than from curated metadata, is an important direction for extending UCVE to domains lacking such structured schemas. Addressing the compounding challenge of sparse rewards in large obligation spaces through reward shaping, curriculum learning or semi-supervised edge initialization represents an important direction for future work. 

We do not evaluate LLM-based query expansion as a baseline. While expanding input queries with generated paraphrases or hypothesized obligations may improve flat retrieval, it is orthogonal to \ac{UCVE}'s graph-based reranking and could be composed with it; we leave this comparison to future work. Under an independently tuned adaptive stopping rule rather than the matched-K protocol, performance differences on CLAUDETTE are not statistically significant, and a heat-kernel diffusion baseline marginally exceeds CRG~+~Spectral's recall on OPP-115 under an unmodified stopping rule (Appendix~\ref{app:adaptive-stopping}).

\bibliography{custom}

\appendix

\section{Graph Construction} \label{app:construction}

\begin{table}[htbp]                                                           
\centering                                   
    \resizebox{\columnwidth}{!}{%
    \begin{tabular}{@{}lccc@{}}                                                           
    \toprule                                             
    Edge Type Removed & OPP-115 $\Delta$R & CLAUDETTE $\Delta$R & ESRS $\Delta$R \\
    \midrule
      \texttt{category} & -0.027 & +0.000 & -0.026 \\
      \texttt{fairness\_tier} & +0.000 & +0.000 & +0.000 \\
      \texttt{kb\_tag} & +0.000 & +0.000 & +0.000 \\
      \texttt{parent\_category} & +0.000 & +0.000 & +0.000 \\
      \texttt{sub\_attribute} & -0.006 & +0.000 & +0.000 \\
      \texttt{framework} & +0.000 & +0.000 & +0.000 \\
      \midrule
      All metadata & -0.034 & -0.012 & -0.026 \\
    \bottomrule
    \end{tabular}
    }
    \caption{Recall change when removing individual metadata edge types.}
    \label{tab:edge-ablation-delta}
\end{table}

\begin{table}[htbp]                 
  \centering
  \begin{tabular}{@{}llcc@{}}
  \toprule
  Dataset & Condition & Recall & Precision \\
  \midrule
  \multirow{4}{*}{OPP}
  & Flat cosine & 0.748 & 0.195 \\
  & Spectral $\times$ Cos. & 0.759 & 0.170 \\
  & BFS $\times$ RL (@ 14) & 0.710 & 0.184 \\
  & Spectral $\times$ RL & \textbf{0.792} & \textbf{0.185} \\
  \midrule
  \multirow{4}{*}{CLA}
  & Flat cosine & 0.806 & 0.122 \\
  & Spectral $\times$ Cos. & 0.848 & 0.117 \\
  & BFS $\times$ RL (@ 16) & 0.806 & 0.113 \\
  & Spectral $\times$ RL & \textbf{0.860} & \textbf{0.118} \\
  \midrule
  \multirow{4}{*}{ESRS}
  & Flat cosine & 0.723 & 0.097 \\
  & Spectral $\times$ Cos. & 0.729 & 0.097 \\
  & BFS $\times$ RL (@ 16) & 0.626 & 0.104 \\
  & Spectral $\times$ RL & \textbf{0.755} & \textbf{0.101} \\
  \midrule
  \multirow{4}{*}{Avg}
  & Flat cosine & 0.735 & 0.138 \\
  & Spectral $\times$ Cos. & 0.779 & 0.128 \\
  & BFS $\times$ RL & 0.655 & 0.150 \\
  & Spectral $\times$ RL & \textbf{0.802} & \textbf{0.135} \\
  \bottomrule
  \end{tabular}
  \caption{Component analysis of graph type $\times$ retrieval method. Each condition uses matched-$K$ while spectral conditions use natural stopping.}
  \label{tab:2x2-ablation}
\end{table}

The \ac{CRG} is initialized using pairwise cosine similarities between \texttt{all-mpnet-base-v2} node embeddings, retaining edges with similarity $\geq 0.5$. This yields dense base graphs—148 edges for OPP-115, 577 for CLAUDETTE, and 595 for ESRS—reflecting high semantic overlap within same-domain requirements.Table~\ref{tab:2x2-ablation} decomposes the contributions of graph type and retrieval method. While spectral reranking on the cosine-only graph yields modest gains, and \ac{BFS} on the \ac{RL}-weighted graph produces inconsistent results, their combination achieves the strongest, most consistent performance. This confirms their conditional dependence: \ac{RL} edge weights reshape the Laplacian's community structure into a form that spectral projection can exploit, whereas neither component succeeds independently. Full edge distributions are reported in Table~\ref{tab:edge-ablation-delta}.

\section{Spectral Community Analysis} \label{app:spectral-communities}
\begin{table}[h]
\centering
\small
\begin{tabularx}{\columnwidth}{@{} l c l X @{}}
\toprule
\textbf{ID} & \textbf{Size} & \textbf{Label} & \textbf{Categories} \\ \midrule
C1 & 7  & Core Practices & data\_retention, data\_security, 1st\_party\_collection, other, 3rd\_party\_sharing, user\_access, user\_choice \\ \addlinespace
C2 & 10 & Collection & First Party Collection/Use sub-attributes (9), Do Not Track (1) \\ \addlinespace
C3 & 2  & Jurisdictional & International and Specific Audiences (2) \\ \addlinespace
C4 & 2  & Policy Meta & Policy Change (1), Other (1) \\ \addlinespace
C5 & 3  & Third-Party & Third Party Sharing/Collection sub-attributes (3) \\ \bottomrule
\end{tabularx}
\caption{OPP-115 Spectral Communities ($k=5, k\text{-NN}=3, n=24$).}
\label{tab:opp_communities}
\end{table}

\begin{table}[h]
\centering
\small
\begin{tabularx}{\columnwidth}{@{} l c l X @{}}
\toprule
\textbf{ID} & \textbf{Size} & \textbf{Label} & \textbf{Categories (Top-1)} \\ \midrule
C1 & 9  & Content Removal & CR (9) --- 100\% \\
C2 & 17 & Liability Limit. & LTD (17) --- 100\% \\
C3 & 5  & Arbitration & A (5) --- 100\% \\
C4 & 15 & Termination & TER (15) --- 100\% \\
C5 & 7  & Unilat. Change & CH (7) --- 100\% \\ \bottomrule
\end{tabularx}
\caption{CLAUDETTE Spectral Communities ($k=5, k\text{-NN}=3, n=53$).}
\label{tab:claudette_communities}
\end{table}

\begin{table}[h]
\centering
\scriptsize %
\setlength{\tabcolsep}{2pt}
\begin{tabularx}{\columnwidth}{@{} l c l X c @{}}
\toprule
\textbf{ID} & \textbf{Sz.} & \textbf{Label} & \textbf{Categories} & \textbf{Pillar} \\ \midrule
C1 & 12 & Gov. \& Strat. & Governance (5), Strategy (5), General (2) & G \\
C2 & 7  & Workforce & Own Workforce (6), Business Conduct (1) & S \\
C3 & 2  & Anti-corrupt. & Business Conduct (2) & G \\
C4 & 26 & Environmental & Climate (9), Pollution (6), Circ. Economy (6), Water (5) & E \\
C5 & 3  & Ethics & Business Conduct (3) & G \\
C6 & 11 & Labor Stds. & Own Workforce (11) & S \\
C7 & 15 & Stakeholder & Value Chain (5), Communities (5), Users (5) & S \\
C8 & 6  & Biodiversity & Biodiversity/Ecosystems (6) & E \\ \bottomrule
\end{tabularx}
\caption{ESRS-82 Spectral Communities ($k=8, k\text{-NN}=5, n=82$).}
\label{tab:esrs_communities}
\end{table}

\begin{figure*}[h]
    \centering
    \includegraphics[scale=0.33]{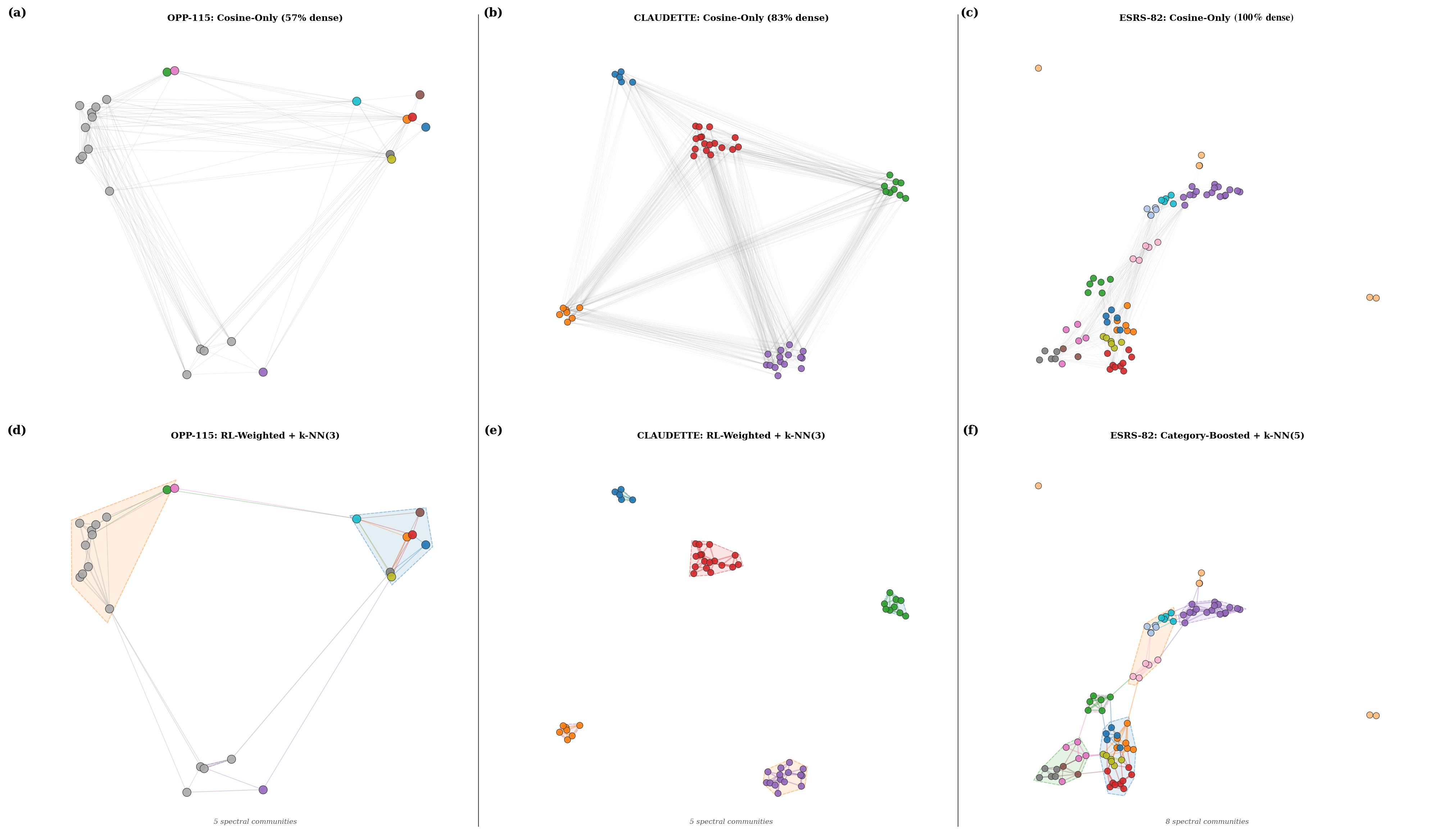}
    \caption{Community structure of the Contextual Requirement Graph across three compliance domains. (a–c) Dense cosine-similarity graphs (57–100\% edge density) where near-complete connectivity obscures meaningful topology. (d–f) After edge weight learning and k-NN sparsification, distinct spectral communities emerge (dashed hulls; 5 eigenvectors for OPP-115/CLAUDETTE, 12 for ESRS-82). Node colors indicate requirement categories. This learned sparse structure is the signal spectral reranking exploits for +3.4–4.5\% recall lift over flat cosine retrieval.}
    \label{fig:community_structure}
\end{figure*}

The spectral communities reveal the qualitative differences in graph structure across each dataset. In CLAUDETTE, community membership aligned perfectly with ground-truth categories, where each of the five unfair clause types --- arbitration, unilateral change, content removal, limitation of liability, and unilateral termination --- forms its own spectral cluster (\ref{tab:claudette_communities}. This can be seen by the cleanly separated clusters by category in Figure \ref{fig:community_structure}. This clean and direct alignment explains why the spectral reranking achieves its largest recall lift. This lift suggests projecting cosine scores onto the Laplacian eigenvectors effectively propagates relevance signal within category boundaries. 

The spectral communities in OPP-115 are affected by the hierarchical structure of the dataset rather than the clean separation present in CLAUDETTE, which is flat. The spectral communities that formed had overlapping datasets (Figure \ref{fig:community_structure}), but upon further analysis, it became apparent each community has specific themes. The largest community (C1) groups together all broad-scope data practice categories  (retention, security, collection, sharing, access, and choice) which occur in virtually every privacy policy (\ref{tab:opp_communities}. These nodes share high embedding similarity (mean cosine 0.635 vs. 0.469 across communities) because their requirement descriptions all follow the pattern "Does the text describe [user data practice]?" The remaining communities (C2–C5) capture more specialized sub-attribute nodes and rare topics (international audiences, policy changes) that are genuinely discriminative. Spectral reranking exploits this by treating C1 as a co-activation group: if one core data practice is relevant, the others likely are too.

ESRS communities partition cleanly along ESG boundaries. The largest cluster (C4) unifies all environmental topics while social requirements are split into three distinct subclusters: general workforce standards (C6), specific labor issues (C2), and external stakeholder impacts (C7). Governance nodes fragment across C1, C3 and C5. This pillar aligned structure enables spectral reranking to boost recall by 3.2\% despite the ESRS subgraph's weaker spectral gap ($\lambda_2 = 0.04$ vs 0.19 for OPP-115)m as the eigenvectors still capture meaningful boundaries. 

\subsection{Structural Signal}
Each \ac{CRG}'s edge types contribute differently across domains, reflecting the structural properties of each dataset. On OPP-115, removing \texttt{same\_category} edges costs $-2.7\%$ recall and removing \texttt{same\_sub\_attribute} edges costs an additional $-0.6\%$. On CLAUDETTE, the three dataset-specific edge types individually contribute negligible recall lift, but removing all three collectively costs $-1.2\%$ recall, suggesting that structural signal in CLAUDETTE is distributed rather than localized to any single edge type. On ESRS, only \texttt{same\_category} contributes meaningfully ($-2.6\%$ recall when removed), with all other edge types contributing zero individually and $-2.6\%$ collectively, mirroring the pattern of OPP-115 but with a single active edge rather than two.  The near-additivity of OPP-115's two active edge types indicates they capture largely independent structural signals; the collective dependence on CLAUDETTE reflects its denser, flatter schema. Full edge type results are reported in Table~\ref{tab:edge-ablation-delta}.

\section{RL Edge Weight Learning Details}
\label{app:rl-details}

We learn a weight vector $\lambda = [\lambda_1, \ldots, \lambda_6]$ over six metadata edge types via epsilon-greedy BFS traversal over the domain graph $G$. The cosine similarity weight is fixed at 1.0 throughout. Training runs for 30 epochs with learning rate $\alpha = 0.05$ and $\epsilon$ decaying from 0.15 to 0.02. Please see Algorithm \ref{alg:rl} for more details regarding the \ac{RL} process. 

\begin{algorithm}[H]
\caption{RL Edge Weight Learning}
\label{alg:rl}
\begin{algorithmic}[1]
\REQUIRE Graph $G$, docs $\mathcal{D}$, budget $B=15$, epochs $T=30$, LR $\alpha$
\ENSURE Learned weight vector $\lambda$

\STATE Initialize $\lambda_t \gets 0.0$ for all metadata types $t$; $\lambda_{\text{cos}} \gets 1.0$

\FOR{epoch $1 \ldots T$}
    \FOR{each document $d \in \mathcal{D}$}
        \STATE $e_q \gets \text{Embed}(d)$
        \STATE Entry node $v_0 \gets \arg\max_{v \in \mathcal{G}} \text{sim}(e_q, e_v)$
        
        \STATE Perform $\epsilon$-greedy BFS for $B$ steps, scoring neighbors $v$ by:
        \STATE $\text{score}(v) = 0.3 \cdot \text{sim}(e_q, e_v) + 0.5 \cdot w(e_{\text{curr} \to v})$
        
        \FOR{each traversed edge $e_k$ at step $k$}
            \STATE Assign credit $c(e_k) = \begin{cases} 
                1.0 - 0.5 \cdot \frac{k}{B} & \text{if dest is GT} \\ 
                +0.1 & \text{if GT within 3 hops} \\ 
                -0.05 & \text{otherwise} 
            \end{cases}$
        \ENDFOR
        
        \FOR{each metadata type $t$}
            \STATE $\Delta\lambda_t \gets \frac{1}{|E_{\text{visited}}|} \sum_{e \in E_{\text{visited}}} c(e) \cdot \frac{\text{weight}_t(e)}{\sum_{t'} \text{weight}_{t'}(e)}$
            \STATE $\lambda_t \gets \text{clamp}(\lambda_t + \alpha \cdot \Delta\lambda_t, [0.0, 3.0])$
        \ENDFOR
    \ENDFOR
\ENDFOR
\RETURN $\lambda$
\end{algorithmic}
\end{algorithm}

\begin{figure*}[h]
    \centering
    \includegraphics[scale=0.40]{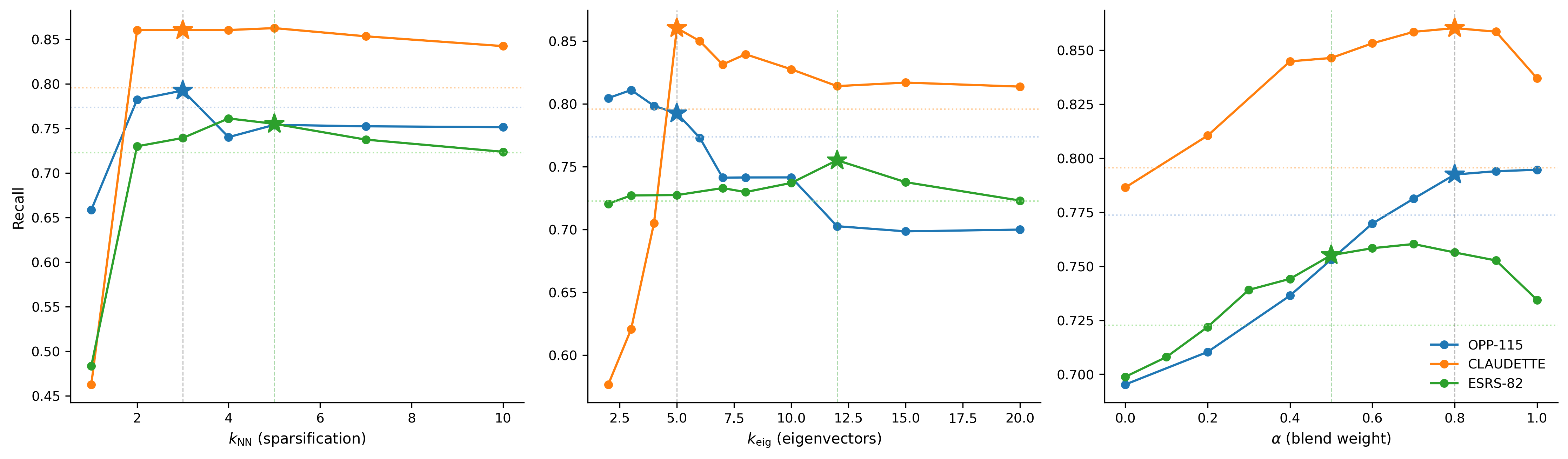}
    \caption{Spectral reranking hyperparameter sensitivity. Each panel sweeps one parameter while holding the others fixed at their chosen values ($k_{\text{NN}}=3$, $k_{\text{eig}}=5$, $\alpha=0.8$). Stars mark chosen values; dotted lines show flat cosine baselines. Performance is stable across wide ranges for $k_{\text{eig}}$ and $\alpha$, but degrades sharply at $k_{\text{NN}}=1$ where over-sparsification disconnects the graph.}
    \label{fig:failed_constraints}
\end{figure*}

\section{Graph Traversal} \label{app:traversal}
\subsection{\ac{GSAEPS}}
Traversal begins from one or more entry nodes selected by the \ac{GSAEPS} procedure. Rather than simply starting from the node with highest cosine similarity to the query --- which may be locally optimal but structurally isolated --- \ac{GSAEPS} scores candidates using a linear combination of features that jointly capture direct relevance, statistical standing within the node set, and structural position in the graph. \\

\begin{table}[h] %
\centering
\small
\setlength{\tabcolsep}{2pt} %
\begin{tabularx}{\columnwidth}{@{} l X cccc @{}}
\toprule
\textbf{Dataset} & \textbf{Entry Strategy} & \textbf{Rec.} & \textbf{Eff.} & \textbf{C70} & \textbf{Vis.} \\
\midrule
\multirow{3}{*}{OPP-115}
  & Random   & 0.967 & 0.160 & 0.725 & 22.6 \\
  & Cosine   & 0.847 & 0.177 & 0.646 & 17.6 \\
  & GSAEPS   & \textbf{0.865} & \textbf{0.174} & \textbf{0.657} & 18.4 \\
\midrule
\multirow{3}{*}{CLA.}
  & Random   & 0.998 & 0.051 & 0.713 & 47.1 \\
  & Cosine   & 0.967 & 0.078 & 0.700 & 30.5 \\
  & GSAEPS   & \textbf{0.967} & 0.078 & 0.700 & 30.6 \\
\midrule
\multirow{3}{*}{Avg.}
  & Random   & 0.982 & 0.106 & 0.719 & 34.9 \\
  & Cosine   & 0.907 & 0.127 & 0.673 & 24.1 \\
  & GSAEPS   & \textbf{0.916} & \textbf{0.126} & \textbf{0.679} & 24.5 \\
\bottomrule
\end{tabularx}

\vspace{4pt}
\begin{minipage}{\columnwidth}
\scriptsize %
\textit{Note:} Random entry achieves highest recall by setting a low sweep threshold, effectively degenerating to flat cosine retrieval. At matched visit budgets (${\sim}18$ nodes), GSAEPS outperforms cosine-only entry by +1.7pp recall on OPP-115.
\end{minipage}
\caption{Effect of entry point strategy on downstream traversal recall.}
\label{tab:entry-ablation}
\end{table}

\subsection{Traversal Strategies}
\textbf{Adaptive Graph \ac{BFS}} performs adaptive-threshold \ac{BFS} from one or more entry nodes, using edge weights to guide neighbor selection at each step. From the current node, all unvisited same-domain neighbors are scored by:
\begin{equation} \label{eq:traversal-score}
    score(v)=0.3 \cdot sim(e_q, e_v) + 0.5 \cdot w(e_{current \rightarrow v})
\end{equation}

where ($e_{current \rightarrow v}$) is the composite \ac{RL}-weighted edge weight from the current node to candidate $v$. This scoring intentionally weighs edge structure more heavily than direct query similarity, encoding the intuition that nodes strongly connected to already-visited relevant nodes are more likely to be relevant than nodes that merely resemble the query text via cosine similarity. Traversal continues as long as the best candidate's similarity exceeds an adaptive stopping threshold:
\begin{equation}
    threshold = max(0.60 \times median(entry_sims), 0.35)
\end{equation}

where $entry\_sims$ is the set of query similarities for all entry nodes. When no remaining candidate exceeds this threshold, traversal halts. This adaptive stopping allows traversal to continue longer when entry nodes are highly similar to the query --- indicating a well-matched document --- and terminate earlier when the match is weaker, naturally scaling the visited set size to document relevance rather than imposing a fixed budget. 

\textbf{Simultaneous Graph Scoring} scores all same-domain nodes simultaneously using a seven dimensional feature vector that combine query similarity and graph structure. For each node $v_i$, the score is:
\begin{equation}
    s(v_i) = w_{gs} \cdot f(v_i)
\end{equation}

where $f(v_i)$ is a seven dimensional feature vector. The features can be found in \ref{tab:graph_score_features}.

\begin{table}[t]
\centering
\small
\renewcommand{\arraystretch}{1.3} %
\begin{tabularx}{\columnwidth}{l X} %
\toprule
\textbf{Feature} & \textbf{Description} \\
\midrule
Query Sim            & Cosine similarity between $e_q$ and $e_i$ \\
Neighbor Avg Wght    & Mean edge weight from $v_i$ to 5 most similar nodes \\
Parent Avg Sim       & Mean query similarity of same-parent nodes \\
Parent Max Sim       & Max query similarity among same-parent nodes \\
Same Parent Top-1    & 1 if $v_i$ shares parent category with top node \\
Query Z-Score        & Z-scored query similarity within dataset \\
Parent Count         & Fraction of dataset nodes in $v_i$'s parent category \\
\bottomrule
\end{tabularx}
\caption{Features used in graph-enhanced node scoring.}
\label{tab:graph_score_features}
\end{table}

All nodes whos score exceeds a fraction of the maximum score is retrieved:
\begin{equation}
    retrieved = \{ v_i : s(v_1) \geq \tau \times max s(v_j)\}
\end{equation}

with $\tau = 0.6$. Simultaneous Graph Scoring does not require sequential traversal decisions or a similarity sweep --- all nodes are scored and filtered in a single pass. Hence, this method is more suited for smaller, more dense graphs where structural signals are informative across the full node set. 

\begin{table*}[t] 
  \centering
  \begin{tabular}{llcccc}
    \toprule
    \textbf{Method} & \textbf{Encoder} & \textbf{OPP-115 R} & \textbf{CLAUDETTE R} & \textbf{ESRS-82 R} & \textbf{Avg R} \\
    \midrule
    \textbf{BM25} & mpnet & 0.649 & 0.732 & 0.490 & 0.624 \\
    \textbf{Flat Cosine} & mpnet & 0.748 & 0.806 & 0.720 & 0.758 \\
    \textbf{PPR} & mpnet & 0.583 & 0.731 & 0.592 & 0.635 \\
    \textbf{Leiden} & mpnet & 0.766 & 0.841 & 0.675 & 0.761 \\
    \textbf{Heat Kernel} & mpnet & 0.747 & 0.817 & 0.744 & 0.769 \\
    \textbf{CRG + Spectral} & \textbf{mpnet} & \textbf{0.792} & \textbf{0.860} & \textbf{0.755} & \textbf{0.802} \\
    \midrule
    \textbf{BM25} & bge-large & 0.641 & 0.732 & 0.492 & 0.622 \\
    \textbf{Flat Cosine} & bge-large & 0.715 & 0.866 & 0.707 & 0.763 \\
    \textbf{PPR} & bge-large & 0.581 & 0.618 & 0.564 & 0.588 \\
    \textbf{Leiden} & bge-large & 0.680 & 0.788 & 0.583 & 0.684 \\
    \textbf{Heat Kernel} & bge-large & 0.715 & 0.866 & 0.718 & 0.766 \\
    \textbf{CRG + Spectral} & \textbf{bge-large} & \textbf{0.758} & \textbf{0.863} & \textbf{0.733} & \textbf{0.785} \\
    \bottomrule
  \end{tabular}
  \caption{Model Performance Evaluation Across Encoders}
  \label{tab:encoder-sensitivity}
\end{table*}

\subsection{Graph-Document Alignment} \label{app: graph-document}

We evaluate a graph-document alignment baseline inspired by GraphCompliance \cite{chung2025graphcompliance}, adapted to our evaluation protocol. The baseline operates in three stages: (1) the document is sentence-segmented and a context graph is constructed by linking sentences sharing key terms above a Jaccard similarity threshold of 0.5; (2) each obligation node's \texttt{validation\_question} is embedded and used to retrieve the top-$k$ most similar sentence nodes from the context graph; (3) an \ac{LLM} alignment gate receives the retrieved sentences and produces a binary compliance verdict per node. Unlike \ac{CRG}, this baseline derives its graph structure from the document rather than from a pre-built obligation space. We note this implementation is an adaptation rather than a faithful reimplementation of GraphCompliance; results should be interpreted accordingly.

\begin{table}[h] %
\centering
\small
\setlength{\tabcolsep}{3pt} %
\begin{tabularx}{\columnwidth}{@{} l l *{5}{R} @{}} %
\toprule
\textbf{Data} & \textbf{Model} & \textbf{Acc} & \textbf{F1} & \textbf{Y-P} & \textbf{Y-R} & \textbf{Y-F1} \\
\midrule
OPP  & 70B & 0.396 & 0.392 & 0.948 & 0.288 & 0.442 \\
CLA  & 8B  & 0.826 & 0.559 & 0.121 & 0.974 & 0.215 \\
ESRS & 8B  & 0.870 & 0.571 & 0.130 & 0.604 & 0.213 \\
\bottomrule
\end{tabularx}
\caption{Graph-document alignment baseline. \textbf{Y-P} and \textbf{Y-R} denote precision/recall over the positive class.}
\label{tab:graph-align}
\end{table}

The results in Table~\ref{tab:graph-align} reveal a precision-recall collapse on CLAUDETTE and ESRS: Yes-recall is high (0.974 and 0.604 respectively) but Yes-precision collapses to 0.121--0.130, indicating that the alignment gate defaults to flagging nearly all nodes as violated. OPP-115 exhibits the inverse pattern, high Yes-precision (0.948) with low Yes-recall (0.288), suggesting that the context graph structure is over-selective in that domain, failing to surface relevant passages for most violated nodes. In both failure modes, the document-derived context graph does not provide sufficient discriminative signal to distinguish violated from satisfied requirements. The failure mode differs by domain: on CLAUDETTE, the alignment gate over-predicts violations, trading recall for near-total loss of precision. On OPP-115, it under-predicts, the inverse trade. \ac{CRG}'s pre-built obligation space and learned edge structure avoids this instability, achieving competitive precision alongside high recall across all three domains without the alignment gate's precision collapse.

\subsection{Adaptive Stopping Robustness Check}
\label{app:adaptive-stopping}

\begin{table}[htbp]
\centering
\resizebox{\columnwidth}{!}{%
\begin{tabular}{lccccc}
\toprule
Method & Rec.@K & $K_{\text{match}}$ & Clamp & Tuned & $\Delta$ \\
\midrule
\multicolumn{6}{l}{\textit{OPP-115 (target $K$=14.19)}} \\
Flat cosine    & .749 & 15.63 & .696 & .746 & $-.037$ {\scriptsize[$-$.057,$-$.016]} \\
Heat kernel    & .754 & 15.41 & .795 & .749 & $-.058$ {\scriptsize[$-$.073,$-$.042]} \\
Cross-encoder  & .780 & 14.48 & .714 & .774 & $-.024$ {\scriptsize[$-$.046,$-$.004]} \\
CRG + Spectral & .791 & --    & .791 & .756 & -- \\
\midrule
\multicolumn{6}{l}{\textit{CLAUDETTE (target $K$=16.42)}} \\
Flat cosine    & .837 & 17.54 & .786 & .827 & $-.027$ {\scriptsize[$-$.072,$+$.014]} \\
Heat kernel    & .855 & 16.82 & .848 & .838 & $-.011$ {\scriptsize[$-$.040,$+$.020]} \\
Cross-encoder  & .814 & 20.59 & .773 & .816 & $-.044$ {\scriptsize[$-$.101,$+$.019]} \\
CRG + Spectral & .860 & --    & .860 & .859 & -- \\
\midrule
\multicolumn{6}{l}{\textit{ESRS (target $K$=16.95, 5-fold mean)}} \\
Flat cosine    & .692 & 20.16 & .686 & .697 & $-.053$ \\
Heat kernel    & .718 & 18.47 & .729 & .718 & $-.016$ \\
Cross-encoder  & .447 & 44.87 & .396 & .435 & $-.287$ \\
CRG + Spectral & .737 & --    & .737 & .719 & -- \\
\bottomrule
\end{tabular} %
}
\caption{Adaptive stopping robustness check. Rec.@K: recall at
CRG~+~Spectral's matched-K budget. $K_{\text{match}}$: average K the
baseline needs to reach CRG~+~Spectral's own recall. Clamp: recall under
the unmodified production stopping rule, no calibration. Tuned: recall at
the independently calibrated operating point. $\Delta$: bootstrap 95\% CI
(OPP-115, CLAUDETTE) or mean difference across folds (ESRS), baseline minus
CRG~+~Spectral at matched K.}
\label{tab:adaptive-stopping}
\end{table}

The matched-K protocol in Table~\ref{tab:main-results} requires all methods
to retrieve the same number of nodes as CRG~+~Spectral's own adaptive
budget. To assess whether a baseline given its own independently tuned
stopping behavior could match CRG~+~Spectral, we compare flat cosine,
heat-kernel diffusion, and a cross-encoder reranker
(\texttt{ms-marco-MiniLM-L-6-v2}) under the production adaptive stopping
rule (Section~\ref{sec:spectral-reranking}), with each baseline's operating
point independently calibrated on a disjoint split to match CRG~+~Spectral's
own average retrieval budget. For ESRS, whose released development split
covers only GRI-standard queries and is therefore not representative of the
ESRS-CID evaluation distribution, we instead calibrate and evaluate via
5-fold cross-validation over source reports.

Table~\ref{tab:adaptive-stopping} reports, for each baseline: recall at
CRG~+~Spectral's matched K, the average K the baseline requires to reach
CRG~+~Spectral's own recall, recall under the unmodified production
stopping rule with no independent calibration (``clamp-only''), and recall
at the tuned operating point. For OPP-115 and CLAUDETTE we additionally
report a paired bootstrap 95\% confidence interval (2,000 resamples over
documents) on the recall difference at matched K; for ESRS we report the
mean difference across folds without a confidence interval.

On OPP-115, CRG~+~Spectral significantly outperforms all three baselines at
matched K (all bootstrap CIs exclude zero). Under the unmodified production
rule with no calibration, however, heat-kernel diffusion attains marginally
higher recall than CRG~+~Spectral's own production point (0.795 vs.\ 0.791);
we report this exception directly, since it uses an identical rule and no
tuning of either method. On CLAUDETTE, differences at matched K are not
significant (all CIs cross zero), indicating comparable performance between
CRG~+~Spectral and the baselines under this protocol. On ESRS, CRG~+~Spectral
outperforms flat cosine (mean $\Delta=-0.053$) and the cross-encoder (mean
$\Delta=-0.287$) by a wide margin; the gap to heat-kernel diffusion is small
(mean $\Delta=-0.016$), and without per-fold confidence intervals we report
this as a numerical rather than statistically confirmed difference.

The clearest pattern is the cross-encoder's degradation with graph size: it
is the strongest baseline on OPP-115 (24 nodes) but requires 44.9 retrieved
nodes on ESRS (82 nodes) to match CRG~+~Spectral's recall at $K{=}17$,
consistent with a pairwise reranker's inability to exploit co-relevance
structure lacking direct lexical overlap.

\section{Encoder Sensitivity} \label{app:encoder-sensitivity}

CRG + Spectral achieves the strongest or tied-strongest recall on OPP-115 and ESRS-82 under both encoders (Table \ref{tab:encoder-sensitivity}). CLAUDETTE shows higher cross-method variance under both encoders, consistent with its dense graph structure (99.8\% edge density) making spectral community separation less stable, a pattern present in the main results as well. The relative gains of CRG + Spectral over flat cosine are consistent across encoders, confirming that the method's contribution is not specific to all-mpnet-base-v2.

\section{Validation}
\subsection{LLM Prompt Templates}

\begin{lstlisting}[caption={Baseline Prompt (\texttt{graph\_context="none"})}, label={lst:prompt-baseline}]
You are a document analyst. You are given a document and a requirement that has been retrieved as a candidate match. The requirement is assumed to be at least somewhat relevant to the document. Your task is to classify whether the document is POSITIVE or NEGATIVE for this requirement.

- Positive means the document text contains language that exhibits, satisfies, or violates this requirement.
- Negative means the document text does NOT contain language that exhibits this requirement, OR the requirement is not applicable to this document.

## Requirement
**Requirement**: Does the report describe the sustainability matters addressed by the governing bodies and the information provided to them for oversight?
**Category**: governance
[...]
\end{lstlisting}

\begin{lstlisting}[caption={Persona Prompt (\texttt{graph\_context="persona"})}, label={lst:prompt-persona}]
[...]
## Context from Graph Traversal
The traversal (5 nodes visited) was connected by same category (100%
Visited requirement topics: "governance", "general".
This requirement was reached because it is connected to "governance" via embedding similarity (weight 0.91).
[...]
\end{lstlisting}

\subsection{Full Pipeline Results - Llama 3.1 8B}
\begin{table}[h]
\centering
\small
\setlength{\tabcolsep}{3pt}
\begin{tabularx}{\columnwidth}{@{} l X rrr @{}}
\toprule
\textbf{Dataset} & \textbf{Strategy} & \textbf{Prec} & \textbf{Rec} &
\textbf{$\Delta$Rec$\uparrow$\textsuperscript{\dag}} \\
\midrule
\multirow{4}{*}{OPP-115}
& Retrieval ceiling & --- & .792 & --- \\
\cmidrule(lr){2-5}
& Baseline & .775 & .564 & $-$.228 \\
& Persona & \textbf{.782} & \textbf{.567} & \textbf{$-$.225} \\
\midrule
\multirow{4}{*}{CLA.}
& Retrieval ceiling & --- & .860 & --- \\
\cmidrule(lr){2-5}
& Baseline & .748 & \textbf{.697} & \textbf{$-$.163} \\
& Persona & \textbf{.773} & .685 & $-$.175 \\
\midrule
\multirow{4}{*}{ESRS}
& Retrieval ceiling & --- & .722 & --- \\
\cmidrule(lr){2-5}
& Baseline & \textbf{.153} & .521 & $-$.201 \\
& Persona & .149 & \textbf{.531} & \textbf{$-$.191} \\
\bottomrule
\end{tabularx}

\vspace{4pt}
\begin{minipage}{\columnwidth}
\scriptsize
\textsuperscript{\dag}$\Delta$Rec$\uparrow$ is the gap between strategy
recall and the retrieval ceiling. Negative values indicate recall below
the ceiling; closer to zero is better. The gap reflects \emph{validation
errors} on retrieved nodes rather than \emph{retrieval failure}.
\end{minipage}

\caption{End-to-end validation performance with graph-derived context
from the \ac{CRG} using Llama-3.1-8B.}
\label{tab:validation-8b}
\end{table}

\subsection{ESRS Per-Pillar Breakdown}
\begin{table}[htbp]
  \centering
  \small
  \setlength{\tabcolsep}{3pt} %
  \begin{tabularx}{\linewidth}{X c cc cc c}
    \toprule
    & & \multicolumn{2}{c}{\textbf{70B}} & \multicolumn{2}{c}{\textbf{8B}} & \\
    \cmidrule(lr){3-4} \cmidrule(lr){5-6}
    \textbf{Pillar} & \textbf{ESG} & \textbf{P} & \textbf{R} & \textbf{P} & \textbf{R} & \textbf{Supp.} \\
    \midrule
    general                 & G & 0.593 & 0.421 & 0.269 & 0.474 & 38 \\
    business conduct        & G & 0.458 & 0.491 & 0.410 & 0.472 & 54 \\
    water marine            & E & 0.467 & 0.250 & 0.315 & 0.304 & 56 \\
    consumers / end users   & S & 0.434 & 0.434 & 0.186 & 0.491 & 53 \\
    value chain workers     & S & 0.405 & 0.464 & 0.218 & 0.603 & 68 \\
    pollution               & E & 0.375 & 0.590 & 0.305 & 0.705 & 61 \\
    own workforce           & S & 0.376 & 0.477 & 0.237 & 0.605 & 86 \\
    governance              & G & 0.282 & 0.312 & 0.150 & 0.594 & 64 \\
    climate change          & E & 0.267 & 0.284 & 0.228 & 0.481 & 81 \\
    biodiversity / ecosyst. & E & 0.254 & 0.485 & 0.200 & 0.515 & 33 \\
    circular economy        & E & 0.217 & 0.574 & 0.165 & 0.672 & 59 \\
    strategy                & G & 0.203 & 0.393 & 0.112 & 0.475 & 61 \\
    affected communities    & S & 0.273 & 0.060 & 0.208 & 0.094 & 51 \\
    \bottomrule
  \end{tabularx}
  \caption{Per-pillar end-to-end precision (P) and recall (R) on ESRS under the baseline validation strategy, for \texttt{Llama-3.3-70B} and \texttt{Llama-3.1-8B}. Support is the approximate number of positive instances per pillar. E/S/G indicates the ESRS pillar category (Environmental, Social, Governance).}
  \label{tab:esrs-pillars}
\end{table}

\end{document}